\documentclass[final,5p,times,twocolumn]{elsarticle}

\usepackage{amssymb}
\usepackage{amsmath}
\usepackage{amsfonts}

\usepackage{graphicx}
\usepackage{booktabs}
\usepackage{tabularx}
\usepackage{multirow}
\usepackage{textcomp}
\usepackage{xcolor}

\usepackage{tcolorbox}
\tcbuselibrary{skins}

\usepackage{url}
\usepackage{hyperref}
\hypersetup{hidelinks}

\usepackage{lineno}

\tcbset{
  problembox/.style={
    colback=gray!5,
    colframe=black!70,
    boxrule=0.5pt,
    arc=2pt,
    left=4pt,right=4pt,top=3pt,bottom=3pt,
    fonttitle=\bfseries\small,
    fontupper=\small,
    title={#1}
  }
}

\tcbset{
  examplebox/.style={
    colback=blue!3,
    colframe=blue!40!black,
    boxrule=0.5pt, arc=2pt,
    left=4pt, right=4pt, top=3pt, bottom=3pt,
    fonttitle=\bfseries\small, fontupper=\small,
    title={#1}
  }
}

\journal{arXiv}
\begin{document}

\begin{frontmatter}
\title{The Logic Chain Breaking Point: A Controlled-Variable Benchmark for Reasoning Degradation in Open-Weight Language Models}
\author[ru]{Akif Islam\corref{cor1}}
\ead{iamakifislam@gmail.com}
\cortext[cor1]{Corresponding author. ORCID: 0009-0001-5300-5310.}
\author[vu]{Mohd. Ruhul Ameen}
\ead{ruhul.cse@vu.edu.bd}
\author[ru]{Md. Ekramul Hamid}
\ead{ekram_hamid@ru.ac.bd}
\affiliation[ru]{organization={Department of Computer Science and Engineering, University of Rajshahi},city={Rajshahi},postcode={6205},country={Bangladesh}}
\affiliation[vu]{organization={Department of Computer Science and Engineering, Varendra University},city={Rajshahi},country={Bangladesh}}
\begin{abstract}
We introduce \textit{The Logic Chain Breaking Point}, a controlled-variable
benchmark for deductive reasoning in open-weight language models. It spans three
phases: (i) pure modus-ponens chain scaling, (ii) distractor resilience
with a fixed core chain, and (iii) contradiction detection under primacy
versus recency placement. Independent variables follow a powers-of-two
sequence $v \in \{1,2,4,8,16,32,64,128\}$, with ground truth known by
construction. We evaluate sixteen open-weight models across
instruction-tuned and reasoning-tuned panels, including a small-model suite
and a larger local-model extension. All
models are evaluated with greedy decoding under model-appropriate context
and output budgets, while context- and length-limited trials are
quarantined from the valid accuracy denominator. Our results reveal a
sharply non-monotonic difficulty surface: pure deductive chains up to
$L{=}128$ remain near saturation for most models, and distractor noise
causes only model-specific degradation at high $D$, while the inconsistent
premise condition has the lowest accuracy. Instruction-tuned models,
including larger 24B--27B baselines, score essentially zero on
inconsistent premise sets, while reasoning-tuned models recover only
partially and remain below reliable accuracy. Within this benchmark, inconsistent premise sets are associated with the
strongest degradation under the benchmark's abstention-based scoring rule.
\end{abstract}
\end{frontmatter}

% Line numbering omitted to keep the two-column review copy readable.

% ----------------------------------------------------------------------
\section{Introduction}
\label{sec:intro}

A language model can often give the right answer. It can solve a word
problem, explain a theorem, write code, or follow a chain of instructions
with impressive fluency. Yet the moment we see a correct answer, a harder
question appears: did the model actually reason its way there, or did it
arrive by matching familiar patterns from its training distribution? This
distinction is not only philosophical. In tasks where the input is
unfamiliar, noisy, or internally inconsistent, a model that relies on
surface patterns may still produce a confident answer, even when the
premises no longer support it. Logical reasoning therefore provides a
useful testbed because the benchmark can apply explicit scoring rules,
including an abstention-based score for inconsistent premises.

This question has become increasingly important as LLMs improve on
reasoning-oriented benchmarks such as GSM8K~\cite{cobbe2021gsm8k},
MATH~\cite{hendrycks2021math}, and BIG-Bench Hard~\cite{suzgun2022bbh}.
High aggregate scores, however, do not always reveal where reasoning
fails. A model may solve a complex-looking problem because it resembles
examples seen during training, while failing on a simpler but more
controlled logical structure. Recent work by Shojaee \emph{et
al.}~\cite{shojaee2025illusion} further suggests that reasoning models can
collapse beyond problem-specific complexity thresholds, even when token
budgets are sufficient. These findings motivate more controlled
evaluations that separate different sources of difficulty rather than
combining them into a single benchmark score.

In this paper, we examine this issue through elementary deductive
reasoning. Instead of using broad word problems or heterogeneous
benchmark items, we construct synthetic logical prompts in which the
ground truth is known by design. The goal is to observe where small
open-weight LLMs remain reliable and where their reasoning begins to
degrade. We ask three connected questions in plain terms: first, how far
can a model follow a pure modus-ponens chain as the number of reasoning
steps increases; second, how much irrelevant but syntactically valid
information can be added before the model loses the relevant chain; and
third, whether the model can recognise an inconsistent premise set,
especially when the contradictory premise appears either before the main
chain or after it.

To answer these questions, we introduce \textit{The Logic Chain Breaking
Point}, a controlled-variable benchmark with three phases. Phase~1 varies
only the length $L$ of a pure deductive chain. Phase~2 keeps the core
chain fixed and varies the number of irrelevant distractor rules $D$.
Phase~3 introduces a contradiction and tests whether models abstain with
\texttt{Indeterminate} rather than forcing a \texttt{True} or
\texttt{False} answer. In all phases, the independent variable follows the
powers-of-two sequence $\{1,2,4,8,16,32,64,128\}$. The benchmark also
separates context and length overflow from genuine reasoning errors, so
that failures caused by inference limits are not counted as logical
failures.

Our results show that the weakest point is not simple chain depth. Across
the evaluated models, pure modus-ponens chains remain largely stable even
at long lengths, and distractors cause only gradual degradation. The
sharpest failure appears when the premise set becomes inconsistent.
Instruction-tuned models tend to follow the valid-looking chain and ignore
the contradiction, while reasoning-tuned models detect the inconsistency
more often but remain far from reliable. This suggests that contradiction
detection is a more diagnostic stressor than chain length alone for
evaluating small open-weight LLMs.

The main contributions of this work are as follows:
\begin{enumerate}
\item We propose a controlled-variable benchmark for analysing deductive
reasoning degradation in small open-weight LLMs, separating chain length,
distractor noise, and contradiction detection into distinct evaluation
phases.
\item We design a reproducible prompt-generation protocol using random
non-semantic entity labels and deterministic seeding, reducing
surface-form leakage and enabling repeatable evaluation.
\item We introduce an overflow-aware evaluation procedure that reports
context- and length-limited trials separately from valid reasoning trials,
avoiding the conflation of inference-budget limits with reasoning failure.
\item We evaluate sixteen open-weight models across instruction-tuned and
reasoning-tuned panels, including both small local models and a larger
RTX-4090/Ollama extension, providing a comparative view of how model type
and size class affect the same logical stressors.
\item We identify contradiction detection as the most fragile condition in
this benchmark, and show that reasoning-tuned models are sensitive to
whether the contradiction is placed before or after the chain, but that
the direction of this placement effect is model-dependent rather than a
single, uniform bias.
\item We report Wilson 95\% confidence intervals alongside every accuracy
figure and table entry, deduplicate and cross-validate all logged trials
against the underlying data-collection code before analysis, and state
sample sizes and early-stopping behaviour explicitly, so that reported
breaking points can be read with their associated uncertainty rather than
as bare point estimates.
\end{enumerate}

% ----------------------------------------------------------------------
\section{Related Work}
\label{sec:related}

\subsection{Reasoning in Large Language Models}

Large language models have shown strong performance on many
reasoning-oriented tasks, especially when they are prompted to produce
intermediate steps before giving the final answer. Chain-of-thought
prompting showed that a model can improve on arithmetic, commonsense, and
symbolic reasoning tasks when examples include step-by-step reasoning
traces~\cite{wei2022cot}. Later work showed that even a simple instruction
such as ``Let's think step by step'' can improve zero-shot reasoning
performance across several tasks~\cite{kojima2022zeroshotcot}.
Self-consistency further improved chain-of-thought prompting by sampling
multiple reasoning paths and selecting the most consistent final
answer~\cite{wang2022selfconsistency}. These studies established that
inference-time reasoning traces can improve final-answer accuracy, but
they mainly evaluate overall task success. They do not directly show which
specific part of a reasoning process fails when the input is made longer,
noisier, or logically inconsistent.

Recent work has also raised caution about interpreting reasoning benchmark
scores as evidence of reliable algorithmic reasoning. Shojaee \emph{et
al.}~\cite{shojaee2025illusion} studied reasoning models using
controllable puzzle environments and found that performance can collapse
beyond certain complexity levels. Their analysis suggests that
final-answer accuracy alone may hide important failure modes, especially
when the benchmark does not isolate the source of difficulty. This
motivates controlled evaluations where one variable is changed at a time
while the logical structure remains fixed.

\subsection{Logical Reasoning Benchmarks}

Several benchmarks have been proposed to evaluate logical reasoning in
natural language. RuleTaker studied whether transformer models can answer
questions from facts and rules requiring different depths of
inference~\cite{clark2020ruletaker}. ProofWriter extended this line of
work by generating implications, proofs, and abductive statements over
natural-language theories~\cite{tafjord2021proofwriter}. These datasets
are important because they connect language understanding with rule-based
deduction and provide explicit proof structures.

Other datasets evaluate broader forms of logical reasoning. LogiQA is
built from expert-written logical reasoning questions and covers different
types of deductive reasoning in reading comprehension~\cite{liu2020logiqa}.
FOLIO provides natural-language premises and conclusions paired with
first-order logic annotations, allowing logical correctness to be checked
with formal tools~\cite{han2022folio}. These benchmarks are valuable for
measuring general logical ability, but they often combine several factors
at once, such as vocabulary, linguistic complexity, rule depth, world
knowledge, and distractor content. As a result, it can be difficult to
identify whether an error is caused by long reasoning depth, irrelevant
information, ambiguity in the wording, or failure to recognise
inconsistency.

\subsection{Distractors, Position Effects, and Inconsistency}

Another related line of work studies whether models can ignore irrelevant
information. Shi \emph{et al.}~\cite{shi2023irrelevant} introduced GSM-IC,
a variant of grade-school math problems with irrelevant context, and
showed that irrelevant information can substantially reduce model accuracy.
This finding is important because real inputs often contain information
that is syntactically valid but not logically useful. However, this work
focuses mainly on arithmetic reasoning, while the present study uses
synthetic propositional chains where distractor rules are logically
disjoint from the target chain.

The position of information inside the prompt can also affect model
behaviour. Liu \emph{et al.}~\cite{liu2023lost} showed that language
models do not always use long contexts robustly and often perform worse
when relevant information appears in less favourable positions. Berglund
\emph{et al.}~\cite{berglund2023reversal} showed that models can fail to
generalise simple relational facts in the reverse direction, even when the
forward relation is learned. These studies suggest that models may rely on
positional and surface-level cues rather than stable symbolic operations.
For contradiction detection, this is especially relevant because the same
inconsistent premise may be treated differently depending on whether it
appears before or after the main chain.

Inconsistency handling remains a difficult evaluation target. A model may
correctly follow a valid chain but still fail to recognise that another
premise contradicts part of that chain. In practical settings such as
retrieval-augmented generation, legal analysis, and scientific question
answering, conflicting evidence is common. Therefore, evaluating whether a
model can detect inconsistency and abstain from a forced binary answer is
an important complement to ordinary reasoning-depth evaluation.

\subsection{Open-Weight Reasoning Models}

Recent open-weight reasoning models have made this question more timely.
DeepSeek-R1 introduced reinforcement-learning-based reasoning models and
released distilled variants based on Qwen and Llama
backbones~\cite{deepseek2025r1}. Phi-4-reasoning and Phi-4-reasoning-plus
were trained to produce detailed reasoning chains and to make better use
of inference-time computation~\cite{abdin2025phi4reasoning}. The Qwen
model family has also reported strong improvements in reasoning, coding,
mathematics, and instruction-following across open-weight
releases~\cite{yang2025qwen3}. These models are increasingly used in local
and resource-constrained settings, but their behaviour under controlled
logical stressors is still not fully understood. In particular, it is
unclear whether reasoning-tuned models are more robust because they
perform more reliable deduction, or whether they mainly improve on
familiar benchmark formats.

\subsection{Research Gap}

Existing work shows that chain-of-thought prompting improves reasoning
performance, that logical reasoning benchmarks can test rule-based
inference, and that irrelevant context or position effects can degrade
model behaviour. However, these strands are usually studied separately.
Many logical benchmarks vary several factors at the same time, making it
difficult to locate the precise source of failure. Distractor studies
often focus on arithmetic word problems rather than formal deductive
chains. Studies of position effects usually examine retrieval or factual
recall rather than contradiction detection inside an explicit logical
premise set.

The gap addressed in this paper is therefore methodological. There is a
need for a simple controlled-variable benchmark that separates three basic
stressors: the length of a pure deductive chain, the amount of irrelevant
but valid-looking rule content, and the presence and placement of a
contradiction. By varying only one factor at a time and using ground truth
known by construction, the proposed benchmark aims to identify not only
whether a model fails, but also what kind of logical pressure causes the
failure.

% ----------------------------------------------------------------------
\section{Methodology}
\label{sec:method}

\subsection{Benchmark Structure}
The benchmark comprises three phases, each designed to isolate a principal
stressor under its specified task conditions. The phases also differ in core
chain length and answer label, so cross-phase accuracy differences are
descriptive and cannot by themselves establish a causal effect of one stressor. Every prompt begins with an
identical instruction preamble directing the model to answer with exactly
one of \texttt{True}, \texttt{False}, or \texttt{Indeterminate}; the
preamble and rules are sent as a single user-turn message, with no
separate system-role message. Entity names below are readable
placeholders (\texttt{X$_i$}); in the actual runs they are random
non-semantic tokens (Section~\ref{sec:prompt}).

\begin{tcolorbox}[problembox={Phase 1 --- Pure Deductive Chain Scaling}]
\textbf{Variable:} chain length $L \in \{1,2,4,8,16,32,64,128\}$.\\
\textbf{Prompt:} a linear modus-ponens chain ``$E_0 \Rightarrow E_1$;
$\dots$; $E_{L-1} \Rightarrow E_L$. Given $E_0$, is $E_L$
\texttt{True}?''.\\
\textbf{Ground truth:} \texttt{True} (chain is fully derivable).\\
\textbf{Tests:} composing a deterministic deductive chain of growing
depth.\\[4pt]
\rule{\linewidth}{0.3pt}\\[2pt]
\textit{\textbf{Worked example ($L{=}3$).}}
Rules: If \texttt{X$_0$} then \texttt{X$_1$}. If \texttt{X$_1$} then
\texttt{X$_2$}. If \texttt{X$_2$} then \texttt{X$_3$}.
Fact: \texttt{X$_0$} holds. Question: Is \texttt{X$_3$} true?\\
Options: \texttt{True} $\;\bullet\;$ \texttt{False} $\;\bullet\;$
\texttt{Indeterminate}.\\
\textbf{Correct:} \texttt{True}
\;(\texttt{X$_0\!\to\!$X$_1\!\to\!$X$_2\!\to\!$X$_3$}).
\end{tcolorbox}

\begin{tcolorbox}[problembox={Phase 2 --- Distractor Resilience}]
\textbf{Variable:} distractor count $D \in \{1,2,4,8,16,32,64,128\}$, core
chain length fixed at $L{=}4$.\\
\textbf{Prompt:} the core 4-step chain interleaved with $D$ syntactically
valid distractor rules, intended to use entities disjoint from the core.\\
\textbf{Ground truth:} \texttt{True}.\\
\textbf{Tests:} ignoring logically inert noise.\\[4pt]
\rule{\linewidth}{0.3pt}\\[2pt]
\textit{\textbf{Worked example ($L{=}4$, $D{=}2$).}}
Core: If \texttt{X$_0$} then \texttt{X$_1$}. $\dots$ If \texttt{X$_3$}
then \texttt{X$_4$}.
Distractors: If \texttt{Y$_0$} then \texttt{Y$_1$}. If \texttt{Z$_0$} then
\texttt{Z$_1$}. \emph{(disjoint, never referenced)}
Fact: \texttt{X$_0$} holds. Question: Is \texttt{X$_4$} true?\\
Options: \texttt{True} $\;\bullet\;$ \texttt{False} $\;\bullet\;$
\texttt{Indeterminate}.\\
\textbf{Correct:} \texttt{True} \;(distractors \texttt{Y}, \texttt{Z} do
not affect the \texttt{X}-chain).
\end{tcolorbox}

\begin{tcolorbox}[problembox={Phase 3 --- Contradiction Detection}]
\textbf{Variables:} placement $C \in \{\text{Primacy}, \text{Recency}\}$,
crossed with distractor count $D \in \{1,2,4,8,16,32,64,128\}$ (the same
sweep as Phase 2); core chain length fixed at $L{=}5$.\\
\textbf{Prompt:} a 5-step chain plus $D$ irrelevant distractor rules, plus
a single rule that directly contradicts an \emph{intermediate} node of the
chain --- the node reached after two hops from the given fact, not the
final queried node --- placed either before (\emph{Primacy}) or after
(\emph{Recency}) the chain and distractors. Contradicting an intermediate
node rather than the queried literal itself prevents the model from
detecting the inconsistency by surface string-matching between the
contradiction and the question; it must instead derive the intermediate
node's truth value via the chain before it can notice the conflict.\\
\textbf{Ground truth:} \texttt{Indeterminate} (premise set is
inconsistent).\\
\textbf{Tests:} recognising an inconsistency that must be propagated
through the chain, and abstaining from deduction.\\[4pt]
\rule{\linewidth}{0.3pt}\\[2pt]
\textit{\textbf{Worked example (Recency placement, $D{=}1$).}}
Chain: If \texttt{X$_0$} then \texttt{X$_1$}. If \texttt{X$_1$} then
\texttt{X$_2$}. If \texttt{X$_2$} then \texttt{X$_3$}. If \texttt{X$_3$}
then \texttt{X$_4$}. If \texttt{X$_4$} then \texttt{X$_5$}.
Distractor: If \texttt{Y$_0$} then \texttt{Y$_1$}. \emph{(disjoint, never
referenced)}
Fact: \texttt{X$_0$} holds. Contradiction (last): \texttt{X$_2$} is false.
Question: Is \texttt{X$_5$} true?\\
Options: \texttt{True} $\;\bullet\;$ \texttt{False} $\;\bullet\;$
\texttt{Indeterminate}.\\
\textbf{Correct:} \texttt{Indeterminate} \;(the chain forces
\texttt{X$_0\!\to\!$X$_1\!\to\!$X$_2$}, so \texttt{X$_2$} is derivably
true, yet a premise directly asserts \texttt{$\neg$X$_2$}: the premise set
is inconsistent, and the benchmark scores an abstention response as correct).
\end{tcolorbox}

\subsection{Interpretation of inconsistency}
The label \texttt{Indeterminate} in Phase~3 is an operational abstention
label assigned by this benchmark when the premises are inconsistent. It is
not a claim that the queried proposition is underdetermined under every
formal logic. In classical logic an inconsistent premise set has no model and,
under the principle of explosion, entails any proposition. The benchmark
instead scores an abstention response as correct for such premises. The
recorded prompt asks for one of the three labels but does not explicitly
instruct the model to use \texttt{Indeterminate} whenever premises conflict.
A model's response may therefore reflect its interpretation of the answer
labels and instructions, not solely its ability to detect a contradiction.

\subsection{Use of generative AI in manuscript preparation}
OpenAI Codex (GPT-6) assisted with language, manuscript formatting, and the
clarification of benchmark limitations during preparation for submission.
The authors are responsible for reviewing the text, data, analysis, and
conclusions.

\subsection{Prompt Construction}
\label{sec:prompt}
The shared prompt opens, ``You are a flawless logical reasoning engine.
Based ONLY on the rules provided below, answer the final question.'' It
then presents the numbered rules, the initial fact and the question, and
ends, ``Provide your final answer on a new line, and only state one of the
following words: True, False, or Indeterminate.'' The parser removes
\texttt{<think>} sections and uses the last standalone answer label in the
remaining response; responses without such a label are parse failures.

All entity labels are randomly drawn non-semantic tokens to prevent
alphabetical-ordering leakage---a confounder identified in earlier
prototypes using \texttt{A, B, C, \ldots} labels. Prompt construction uses
a deterministic character-rolling hash rather than Python's built-in
\texttt{hash()}, ensuring byte-identical prompts across resumed runs
irrespective of \texttt{PYTHONHASHSEED}. Because the random-number
generator is seeded from the (phase, $v$, placement, trial) tuple alone
and not from the model identifier, every model is evaluated on the
identical set of generated prompts, which removes prompt sampling as a
confound in cross-model comparisons. Each (phase, $v$, placement)
condition is evaluated over 20 trials with distinct seeds; for Phase 3,
which crosses two placement variants, this yields 40 trials per (phase,
$v$) cell.

\subsection{Models}

We evaluate sixteen open-weight language models divided by model type and
size class. The first axis separates instruction-tuned models from
reasoning-oriented models. Instruction-tuned models are expected to
provide short direct answers, whereas reasoning-oriented models may produce
longer intermediate reasoning traces before the final answer. The second
axis separates the original small-model panel from a larger local-model
extension evaluated on a 24~GB RTX~4090.

The small instruction-tuned panel contains five models:
\texttt{google.gemma-4-12b-it}~\cite{gemmateam2026gemma4},
\texttt{meta-llama-llama-3.1-8b-instruct}~\cite{grattafiori2024llama3},
\texttt{mistralai\_-\_mistral-7b-instruct-v0.3}~\cite{jiang2023mistral7b},
\texttt{microsoft\_phi-4-mini-instruct}~\cite{abdin2025phi4mini,abdin2024phi4}, and
\texttt{granite-3.3-8b-instruct}~\cite{granite2025ibm33,granite2024ibm}. The larger instruction-tuned panel
contains two models: \texttt{mistral-small3.2:24b}~\cite{mistralai2025smallmodelcard} and
\texttt{gemma3:27b-it-qat}~\cite{gemmateam2025gemma3}.

The small reasoning-oriented panel contains six models:
\texttt{qwen/qwen3.5-9b}~\cite{qwen2026qwen35,yang2025qwen3}, \texttt{google/gemma-4-12b-qat}~\cite{gemmateam2026gemma4},
\texttt{google/gemma-4-e2b}~\cite{gemmateam2026gemma4},
\texttt{deepseek/deepseek-r1-distill-qwen-7b}~\cite{deepseek2025r1},
\texttt{deepseek/deepseek-r1-distill-llama-8b}~\cite{deepseek2025r1}, and
\texttt{microsoft/phi-4-reasoning-plus}~\cite{abdin2025phi4reasoning}. The larger reasoning-oriented
panel contains three models: \texttt{gpt-oss:20b}~\cite{openai2025gptoss},
\texttt{deepseek-r1:14b}~\cite{deepseek2025r1}, and \texttt{phi4-reasoning:plus}~\cite{abdin2025phi4reasoning}.
For brevity, the remainder of the paper drops vendor path prefixes when
referring to individual models (e.g.\ \texttt{gemma-4-e2b} for
\texttt{google/gemma-4-e2b}, \texttt{phi-4-reasoning-plus} for
\texttt{microsoft/phi-4-reasoning-plus}); Table~\ref{tab:model_config_all}
and this section give the full identifiers.

\subsection{Infrastructure and Decoding}

Inference was performed locally using OpenAI-compatible servers. The
original small-model experiments were run through LM Studio on a single
RTX~3060~Ti GPU with 8~GB VRAM. The larger local-model extension was run
through Ollama on an RTX~4090 GPU with 24~GB VRAM. All experiments used
greedy decoding with \texttt{temperature}=0.

Output budgets were set according to model type. Instruction-tuned models
were evaluated with a short completion budget because the required output
is one of three labels: \texttt{True}, \texttt{False}, or
\texttt{Indeterminate}. Reasoning-oriented models were evaluated with
larger completion budgets because they may emit longer reasoning traces
before the final answer. Table~\ref{tab:model_config_all} reports the
context length, maximum output budget, model type, and size class used
for each model.

We explicitly separate inference-budget failures from reasoning failures.
Trials are marked as \texttt{CONTEXT\_LIMIT} when the prompt exceeds the
configured context window, and as \texttt{LENGTH\_LIMIT} when the model
output is truncated by the completion budget. These trials are excluded
from the valid accuracy denominator and reported separately.

\begin{table*}[!t]
\caption{Model configuration used in the benchmark. Context length and
maximum output length are reported in tokens. Type indicates whether the
model is instruction-tuned (short direct answers expected) or
reasoning-oriented (may emit an intermediate chain-of-thought trace).}
\label{tab:model_config_all}
\centering
\footnotesize
\begin{tabular}{@{}p{0.42\textwidth}rrcc@{}}
\toprule
\textbf{Model} & \textbf{Context} & \textbf{Max out} & \textbf{Type} &
\textbf{Size} \\
\midrule
\texttt{mistral-small3.2:24b} & 8192 & 128 & Instruct & Large \\
\texttt{gemma3:27b-it-qat} & 8192 & 128 & Instruct & Large \\
\texttt{gpt-oss:20b} & 32768 & 8192 & Reasoning & Large \\
\texttt{deepseek-r1:14b} & 32768 & 8192 & Reasoning & Large \\
\texttt{phi4-reasoning:plus} & 32768 & 8192 & Reasoning & Large \\
\midrule
\texttt{google.gemma-4-12b-it} & 8192 & 128 & Instruct & Small \\
\texttt{meta-llama-llama-3.1-8b-instruct} & 8192 & 128 & Instruct &
Small \\
\texttt{mistralai\_-\_mistral-7b-instruct-v0.3} & 8192 & 128 & Instruct &
Small \\
\texttt{microsoft\_phi-4-mini-instruct} & 8192 & 128 & Instruct &
Small \\
\texttt{granite-3.3-8b-instruct} & 8192 & 128 & Instruct & Small \\
\texttt{qwen/qwen3.5-9b} & 32768 & 8192 & Reasoning & Small \\
\texttt{google/gemma-4-12b-qat} & 16384 & 6144 & Reasoning & Small \\
\texttt{google/gemma-4-e2b} & 16384 & 6144 & Reasoning & Small \\
\texttt{deepseek/deepseek-r1-distill-qwen-7b} & 32768 & 8192 & Reasoning &
Small \\
\texttt{deepseek/deepseek-r1-distill-llama-8b} & 32768 & 8192 & Reasoning &
Small \\
\texttt{microsoft/phi-4-reasoning-plus} & 8192 & 8192 & Reasoning &
Small \\
\bottomrule
\end{tabular}
\end{table*}

\subsection{Overflow Quarantine}
Despite the margins in Table~\ref{tab:model_config_all}, a trial is
classified \texttt{CONTEXT\_LIMIT} if the prompt exceeds the model's window
or \texttt{LENGTH\_LIMIT} if the completion is cut by \texttt{max\_tokens}.
Such trials are \emph{excluded} from the accuracy denominator and reported
separately in Fig.~\ref{fig:status_composition}. Counting overflow as
failure would conflate model capability with hardware budget.

\subsection{Early Stopping}
\label{sec:earlystop}
For a given model and phase, values of $v$ (chain length $L$ in Phase~1;
distractor count $D$ in Phases~2 and~3) are tested in increasing order.
After each value's 20 (or, for Phase~3, 40) trials complete, if the
valid-trial accuracy at that value is below $0.95$, or if every trial at
that value was quarantined as \texttt{CONTEXT\_LIMIT}, the run stops and
larger values of $v$ are not tested for that (model, phase) pair. This
early-stopping rule was applied to the original small-model suite (both
instruction-tuned and reasoning-tuned panels) and to the larger
reasoning-tuned extension. It was disabled for the larger
instruction-tuned extension (\texttt{mistral-small3.2:24b} and
\texttt{gemma3:27b-it-qat}), which was instead run to completion across
the full $v \in \{1,\dots,128\}$ sweep in every phase. This asymmetry
tracks per-trial compute cost rather than being an oversight: as
Table~\ref{tab:model_config_all} shows, instruction-tuned models are
capped at a 128-token completion budget, so sweeping all eight values of
$v$ across both placements in every phase for two such models is cheap
even without early stopping; reasoning-oriented models are budgeted up to
8192 output tokens and routinely use much of that budget generating an
explicit reasoning trace, making an un-truncated full sweep across the
larger reasoning panel prohibitively slow on the same RTX~4090. Running
the large instruction-tuned panel to completion was therefore both
affordable and useful: because these models score at or near zero on
Phase~3 from the first condition onward, only the full sweep reveals
whether that failure ever relents at higher distractor counts (it
does not, except for a single trial; Section~\ref{sec:results}). This is
why Fig.~3 and Table~\ref{tab:breakpoints} can report a full Phase~3
distractor sweep only for these two models, while every other model is
observed at a single Phase~3 condition ($D=1$) because it already falls
below the threshold there. Early stopping means that a model's absence
from higher values of $v$ in the figures reflects this stopping rule
rather than a missing measurement.

\subsection{Metrics}
For each (model, phase, $v$) cell we compute the valid-trial accuracy: the
fraction of non-quarantined trials with a correct answer, out of 20 trials
(40 for Phase~3). A model's \emph{breaking point} at a phase is the
smallest tested $v$ at which this valid-trial accuracy first drops below
the $0.95$ reliability threshold used for early stopping
(Section~\ref{sec:earlystop}); this is a point-estimate criterion on a
fixed number of trials rather than a confidence-interval bound, and we
report it as such throughout. With $n=20$ trials per cell, a single
incorrect trial moves the observed accuracy from $1.000$ to $0.950$; we
flag such boundary cases explicitly in the text where they occur rather
than treating $0.950$ as unambiguously ``passing''.

To make the uncertainty at these trial counts explicit, we additionally
compute the Wilson score 95\% confidence interval~\cite{wilson1927} for
every reported proportion. These intervals are shown as error whiskers on
Figs.~\ref{fig:p1_all_models}--\ref{fig:p3_all_models} and reported
alongside the point estimate in Table~\ref{tab:breakpoints}; they are a
descriptive uncertainty band around each observed proportion; the
early-stopping and breaking-point decisions themselves remain governed by
the point-estimate rule above; we do not re-derive them from the CI.

% ----------------------------------------------------------------------
\section{Results and Discussion}
\label{sec:results}

\subsection{Phase 1: Chain Length Scaling}

\begin{figure*}[!t]
\centering
\includegraphics[width=0.86\textwidth]{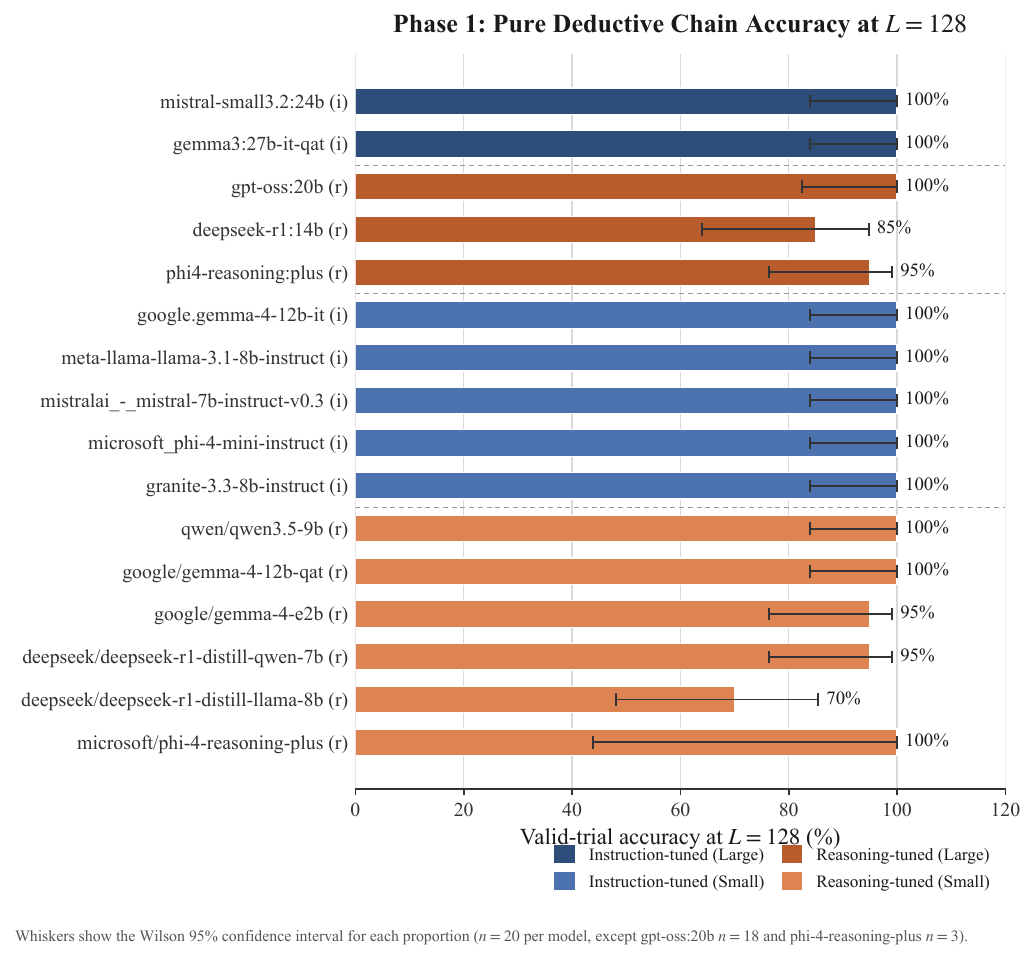}
\caption{Phase 1 valid-trial accuracy at $L=128$ across all sixteen
models, grouped by size class (Large/Small) and model type; $(i)$ denotes
instruction-tuned models and $(r)$ denotes reasoning-oriented models.
Dashed rules separate the four panels: large instruction-tuned, large
reasoning-tuned, small instruction-tuned, and small reasoning-tuned.}
\label{fig:p1_all_models}
\end{figure*}

Fig.~\ref{fig:p1_all_models} shows Phase 1 accuracy at the longest tested
chain length, $L=128$, across all sixteen models. This phase is the
cleanest test in the benchmark because there are no distractors and no
contradictions. The model only needs to follow a sequence of implications
from the starting fact to the final target.

The main observation is that chain length alone does not create an early
failure point. Across both panels, most models remain at or near perfect
accuracy from $L=1$ through $L=64$. Degradation appears mainly at $L=128$
and only for a subset of the reasoning models. At this longest chain
length, deepseek-r1-distill-llama-8b reaches 0.700 accuracy (95\% CI
[0.48, 0.86]), the only small reasoning model to clearly cross the
reliability threshold at this length. Two further small reasoning models
land exactly on the threshold: deepseek-r1-distill-qwen-7b and gemma-4-e2b
both reach 0.950 accuracy (19 of 20 trials correct; 95\% CI [0.76, 0.99]),
one incorrect trial away from breaking --- the width of this interval at
$n=20$ is itself a reason to read the 0.95 cutoff as indicative rather
than exact. In contrast, all five instruction-tuned models remain at
1.000 accuracy at $L=128$.

This result is important because it weakens the simple assumption that
small models fail mainly because the chain becomes long. In this
controlled setting, a linear modus-ponens chain over random symbols
remains manageable for most models, even at 128 steps. The task is not
trivial, but it is structurally regular: each rule points to the next
rule, and there is no competing information. Under those conditions, both
reasoning-tuned and instruction-tuned models can usually preserve the
chain.

The larger-model extension confirms the same pattern. Both large
instruction-tuned models, \texttt{mistral-small3.2:24b} and
\texttt{gemma3:27b-it-qat}, maintain 1.000 accuracy through $L{=}128$.
Among the larger reasoning models, \texttt{gpt-oss:20b} also remains at
1.000 valid-trial accuracy at $L{=}128$, with two length-limited trials
quarantined from the denominator. \texttt{phi4-reasoning:plus} reaches
0.950 at $L{=}128$, while \texttt{deepseek-r1:14b} drops to 0.850 at the
same chain length. Thus, adding larger local models does not overturn the
main Phase 1 conclusion: clean linear chain depth is not the dominant
source of failure.

\subsection{Phase 2: Distractor Resilience}

\begin{figure*}[!t]
\centering
\includegraphics[width=0.86\textwidth]{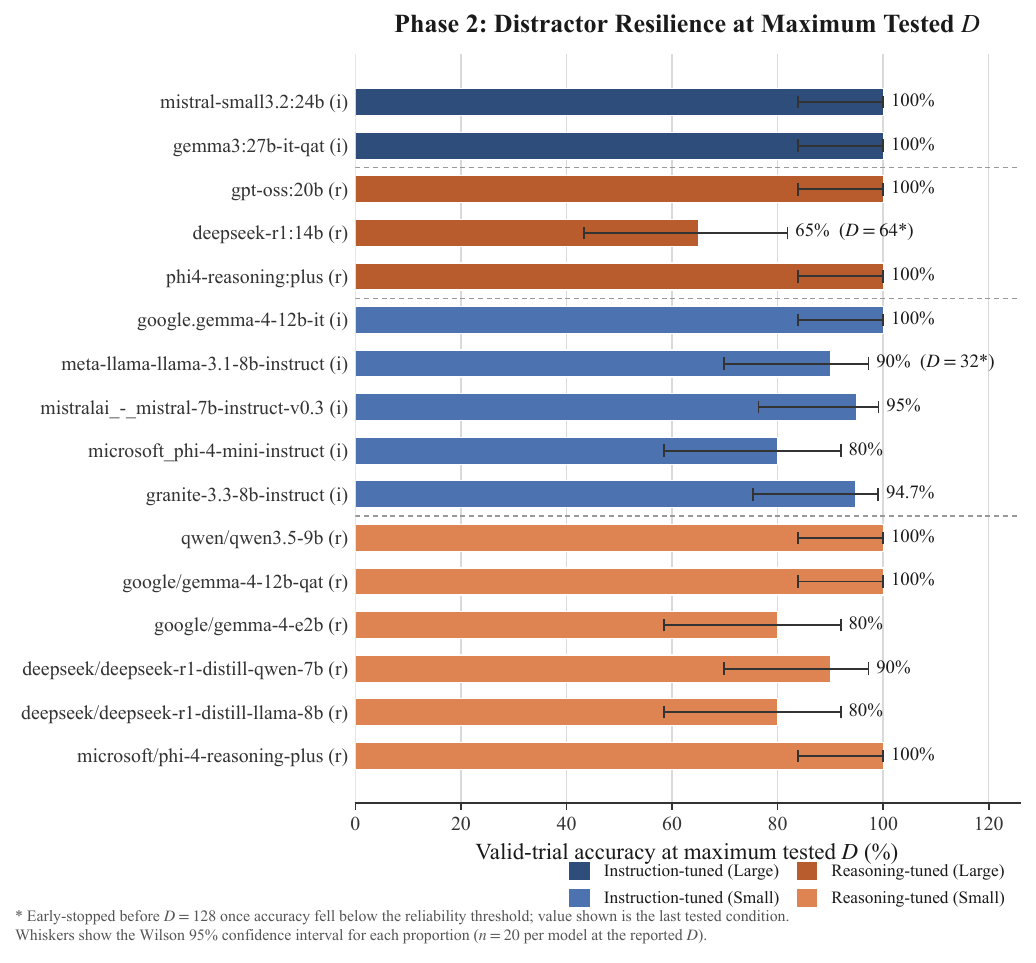}
\caption{Phase 2 valid-trial accuracy at the maximum tested distractor
count across all sixteen models. Values are reported at $D=128$ unless
early stopping occurred; early-stopped models
(\texttt{meta-llama-llama-3.1-8b-instruct} at $D=32$ and
\texttt{deepseek-r1:14b} at $D=64$) are annotated with an asterisk and
their final tested $D$.}
\label{fig:p2_all_models}
\end{figure*}

Fig.~\ref{fig:p2_all_models} shows Phase 2 accuracy at the maximum tested
distractor count for each model while keeping the core deductive chain
fixed at $L=4$. This phase asks a different question from Phase 1. The
model no longer needs to follow a very long chain, but it must identify
which rules matter and which rules can be ignored.

Most models remain stable through $D=32$ distractors. The first clear drop
appears for meta-llama-llama-3.1-8b-instruct, which reaches 0.900 accuracy
at $D=32$. At $D=128$, the weakest results are observed for gemma-4-e2b at
0.800, microsoft\_phi-4-mini-instruct at 0.800, and
deepseek-r1-distill-llama-8b at 0.800. In contrast, phi-4-reasoning-plus
and qwen3.5-9b retain 1.000 accuracy through $D=128$.

The pattern in Phase 2 is gradual rather than abrupt. Adding irrelevant
rules makes the prompt more crowded, but it does not immediately break the
models. This supports the view that distractor noise is a real stressor,
but not the most severe one in this benchmark. It also shows that
reasoning-tuned models do not uniformly dominate instruction-tuned
baselines on distractor resilience. Some instruction models remain highly
reliable even with many irrelevant rules, while some reasoning models begin
to degrade at the highest distractor levels.

The larger instruction-tuned models also remain robust to distractor
noise. Both \texttt{mistral-small3.2:24b} and \texttt{gemma3:27b-it-qat}
achieve 1.000 accuracy through $D{=}128$. The larger reasoning models are
more mixed. \texttt{gpt-oss:20b} and \texttt{phi4-reasoning:plus} remain at
1.000 accuracy through $D{=}128$, whereas \texttt{deepseek-r1:14b} drops to
0.650 at $D{=}64$, triggering early stopping before $D{=}128$. This shows
that distractor resilience does not improve uniformly with model size or
reasoning orientation; it remains model-specific.

\subsection{Phase 3: Contradiction Detection}

\begin{figure*}[!t]
\centering
\includegraphics[width=0.86\textwidth]{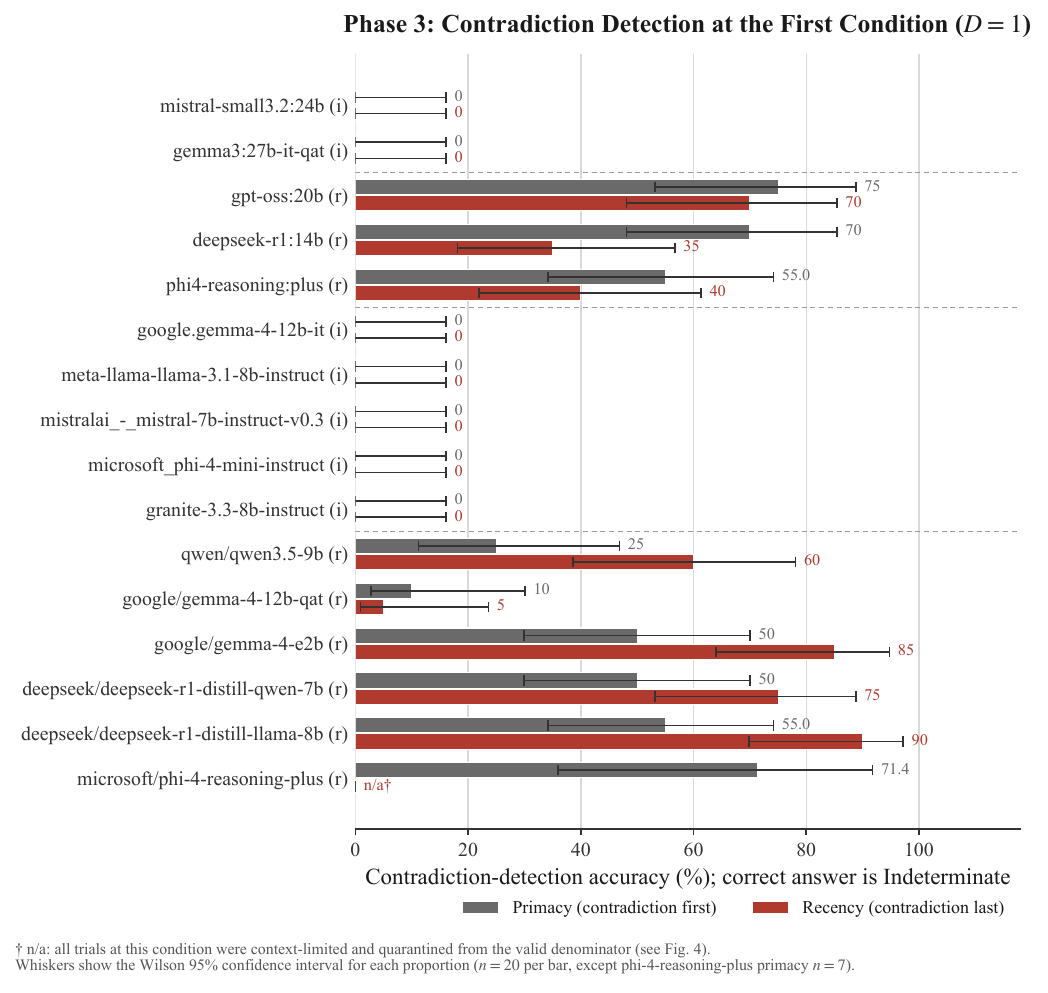}
\caption{Phase 3 contradiction-detection accuracy across all sixteen
models at the first contradiction condition ($D=1$), split by placement:
primacy (contradiction stated before the chain) versus recency
(contradiction stated after the chain). The correct answer is
\texttt{Indeterminate}. \texttt{microsoft/phi-4-reasoning-plus} has no
valid recency trials at this condition (all 20 were context-limited and
quarantined; see Fig.~\ref{fig:status_composition}).}
\label{fig:p3_all_models}
\end{figure*}

Fig.~\ref{fig:p3_all_models} shows that Phase 3 gives the clearest
separation between the model groups. In this phase, the model must do more
than follow the chain. It must notice that the premise set is internally
inconsistent and avoid forcing an ordinary True or False answer.

The instruction-tuned panel fails consistently on this phase. All five
instruction-tuned models score 0.000 on contradiction detection for both
primacy and recency placement. Their responses usually follow the
valid-looking chain and return True, even though the prompt contains a
contradictory premise. This suggests that the models treat the task mainly
as ordinary chain completion rather than inconsistency detection.

The reasoning panel performs better, but the improvement is partial. None
of the reasoning-tuned models reaches the 0.95 threshold on contradiction
detection. When the contradiction is placed at the end of the prompt,
deepseek-r1-distill-llama-8b reaches 0.900, gemma-4-e2b reaches 0.850,
deepseek-r1-distill-qwen-7b reaches 0.750, qwen3.5-9b reaches 0.600, and
gemma-4-12b-qat reaches 0.050. When the contradiction is placed at the
beginning of the prompt, phi-4-reasoning-plus reaches 0.714,
deepseek-r1-distill-llama-8b reaches 0.550, deepseek-r1-distill-qwen-7b
and gemma-4-e2b each reach 0.500, qwen3.5-9b reaches 0.250, and
gemma-4-12b-qat reaches 0.100.

This phase changes the interpretation of the benchmark. The models that
handle long clean chains and many distractors can still fail when the
premise set becomes inconsistent. The placement effect is also important,
although its direction is not universal. For four of the five small
reasoning models with a valid primacy--recency comparison
(gemma-4-e2b, deepseek-r1-distill-llama-8b, deepseek-r1-distill-qwen-7b,
and qwen3.5-9b), a contradiction placed near the end is easier to detect
than one placed near the beginning, consistent with a recency-sensitive
behaviour: the model is more likely to respect the contradiction when it
appears close to the final question, and more likely to overlook it when
it appears before the main chain. \texttt{gemma-4-12b-qat} is the
exception within this panel: it scores higher under primacy (0.100) than
recency (0.050), the opposite direction. The larger-model extension,
discussed next, weakens any claim of a single, universal recency bias
further.

The larger-model extension strengthens the central Phase 3 finding. The
two larger instruction-tuned models still fail almost completely on
contradiction detection. \texttt{gemma3:27b-it-qat} obtains 0.000 accuracy
across the Phase 3 sweep, while \texttt{mistral-small3.2:24b} obtains only
0.003 overall accuracy, with a small non-zero value appearing only at the
largest distractor setting. Therefore, increasing instruction-model size
into the 24B--27B range does not solve inconsistency recognition in this
benchmark.

The larger reasoning models recover more often, but remain far below
reliable performance. At the first contradiction condition,
\texttt{gpt-oss:20b} reaches 0.725 accuracy (95\% CI [0.57, 0.84]),
\texttt{deepseek-r1:14b} reaches 0.525 (95\% CI [0.38, 0.67]), and
\texttt{phi4-reasoning:plus} reaches 0.475 (95\% CI [0.33, 0.63]). None
reaches the 0.95 reliability threshold, and none of these three intervals
excludes the others, so this panel should not be over-read as a strict
ranking. The placement pattern is also not uniformly recency-favouring in
the larger reasoning panel: \texttt{gpt-oss:20b} performs similarly on
primacy (0.750, CI [0.53, 0.89]) and recency (0.700, CI [0.48, 0.86]) ---
the two intervals overlap almost entirely --- while \texttt{deepseek-r1:14b}
and \texttt{phi4-reasoning:plus} perform better when the contradiction
appears first. This suggests that contradiction detection is not merely a
scale problem; it depends strongly on model training and answer-selection
behaviour.

\subsection{Response Composition}
\label{sec:respcomp}

\begin{figure*}[!t]
\centering
\includegraphics[width=0.86\textwidth]{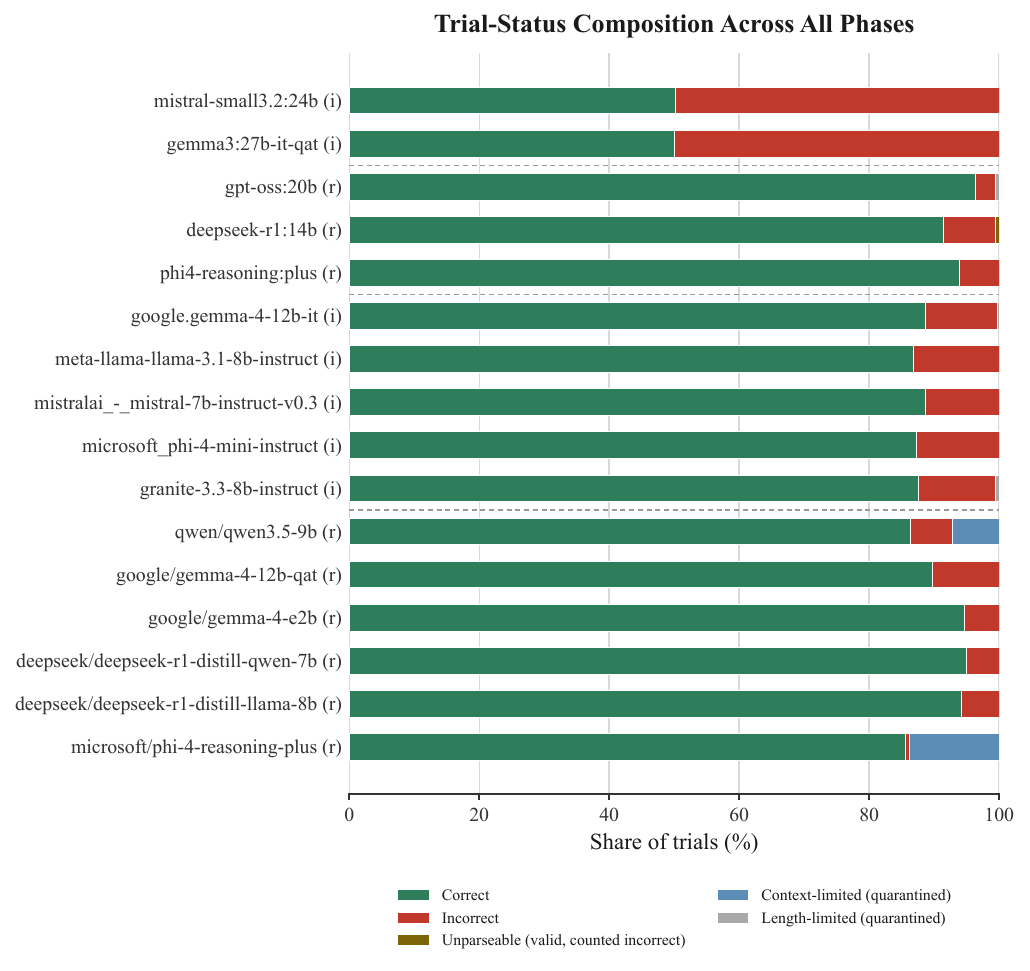}
\caption{Trial-status composition across all sixteen models, pooled over
all three phases. Green denotes correct answers, red denotes incorrect
answers (including the two parse failures for \texttt{deepseek-r1:14b},
shown in olive), and blue/grey denote context- and length-limited trials
that are quarantined from the valid accuracy denominator.}
\label{fig:status_composition}
\end{figure*}

Fig.~\ref{fig:status_composition} shows the distribution of correct,
incorrect, and overflow-quarantined trials across all sixteen models and
all three phases combined. This view is useful because it separates
actual reasoning errors from trials where the answer could not be fairly
evaluated due to context or length limits.

The response-composition plots show that overflow is not the main
explanation for the results. Some context-limited cases appear in the
reasoning panel, especially for models constrained by practical loading
settings on the 8~GB GPU. These trials are quarantined and excluded from
the accuracy denominator. This prevents long-output reasoning models from
being unfairly penalised when the response is truncated or the prompt
exceeds the available context.

For the instruction panel, overflow is essentially absent. The main source
of non-correct responses is ordinary incorrect prediction, concentrated
almost entirely in Phase 3. This supports the central result: the failure
in contradiction detection is not caused by token limits, but by the
model's answer behaviour under inconsistent premises.

In the larger-model extension, overflow remains limited. The larger
instruction-tuned models complete all trials without context or length
limits. In the larger reasoning panel, \texttt{gpt-oss:20b} has two
length-limited trials at $L{=}128$, which are quarantined from the valid
accuracy denominator. \texttt{deepseek-r1:14b} produces two parse failures,
one at $L{=}128$ and one at $D{=}64$, which are counted as valid incorrect
trials. These cases do not explain the main contradiction-detection result,
which remains dominated by ordinary incorrect answers rather than
infrastructure limits.

\subsection{Summary of Breaking Points}

Table~\ref{tab:breakpoints} summarises the model-level results across the
three phases. For Phase 1 and Phase 2, the table reports the first tested
value where the model falls below the target accuracy threshold, or NR if
the threshold is not crossed by the largest tested value. For Phase 3, the
table reports contradiction-detection accuracy directly because the main
condition is inconsistency recognition rather than increasing chain length
or distractor count.

\begin{table*}[!t]
\caption{Per-model breaking points across phases. NR indicates that
valid-trial accuracy never dropped strictly below $0.95$ up to
$v_{\max}=128$; two Phase~1 entries reported as NR
(\texttt{deepseek-r1-distill-qwen-7b} and \texttt{gemma-4-e2b}) reach
exactly $0.950$ at $L=128$, a single trial away from the threshold (see
Section~\ref{sec:results}). Phase 3 reports contradiction-detection
accuracy with its Wilson 95\% confidence interval in brackets; $n=40$
per model except \texttt{phi-4-reasoning-plus} ($n=7$;
Section~\ref{sec:respcomp}) and the two large instruction-tuned models
($n=320$; Section~\ref{sec:earlystop}).}
\label{tab:breakpoints}
\centering
\footnotesize
\renewcommand{\arraystretch}{1.12}
\begin{tabularx}{\textwidth}{@{}>{\raggedright\arraybackslash}Xcccc@{}}
\toprule
\textbf{Model} & \textbf{P1 ($L$)} & \textbf{P2 ($D$)} &
\textbf{P3 Acc.\ [95\% CI]} & \textbf{Size} \\
\midrule
\multicolumn{5}{@{}l}{\textit{Reasoning models}} \\
\texttt{gpt-oss:20b} & NR & NR & 0.725 [0.57, 0.84] & Large \\
\texttt{deepseek-r1:14b} & 128 & 64 & 0.525 [0.38, 0.67] & Large \\
\texttt{phi4-reasoning:plus} & NR & NR & 0.475 [0.33, 0.63] & Large \\
\texttt{deepseek-r1-distill-llama-8b} & 128 & 128 & 0.725 [0.57, 0.84] & Small \\
\texttt{deepseek-r1-distill-qwen-7b} & NR & 128 & 0.625 [0.47, 0.76] & Small \\
\texttt{gemma-4-12b-qat} & NR & NR & 0.075 [0.03, 0.20] & Small \\
\texttt{gemma-4-e2b} & NR & 128 & 0.675 [0.52, 0.80] & Small \\
\texttt{phi-4-reasoning-plus} & NR & NR & 0.714 [0.36, 0.92] & Small \\
\texttt{qwen3.5-9b} & NR & NR & 0.425 [0.29, 0.58] & Small \\
\midrule
\multicolumn{5}{@{}l}{\textit{Instruction-tuned models}} \\
\texttt{mistral-small3.2:24b} & NR & NR & 0.003 [0.00, 0.02] & Large \\
\texttt{gemma3:27b-it-qat} & NR & NR & 0.000 [0.00, 0.01] & Large \\
\texttt{google.gemma-4-12b-it} & NR & NR & 0.000 [0.00, 0.09] & Small \\
\texttt{granite-3.3-8b-instruct} & NR & 128 & 0.000 [0.00, 0.09] & Small \\
\texttt{meta-llama-llama-3.1-8b-instruct} & NR & 32 & 0.000 [0.00, 0.09] & Small \\
\texttt{microsoft\_phi-4-mini-instruct} & NR & 128 & 0.000 [0.00, 0.09] & Small \\
\texttt{mistralai\_-\_mistral-7b-instruct-v0.3} & NR & NR & 0.000 [0.00, 0.09] & Small \\
\bottomrule
\end{tabularx}
\end{table*}

The table reinforces the same pattern seen in the figures. Chain length
and distractor count create some degradation, but the failures are limited
and model-dependent. Contradiction detection is different: it separates the
two model panels sharply and remains difficult even for reasoning-tuned
models. In this benchmark, inconsistent premises are associated with lower
accuracy under the benchmark's abstention-based scoring rule than clean chains are under
their own scoring rule.

% ----------------------------------------------------------------------
\section{Conclusion and Future Work}
\label{sec:conclusion}

We presented a controlled-variable benchmark that disentangles three
stressors on deductive reasoning in open-weight language models: linear
chain depth, irrelevant distractor rules, and contradiction detection.
Across sixteen models, including both small local models and a larger
RTX-4090/Ollama extension, the results show that linear deductive depth is
not the main binding constraint. Most models handle pure modus-ponens
chains up to $L{=}128$ with near-saturation accuracy when no distractors or
contradictions are present. Throughout, we report these accuracies with
their Wilson 95\% confidence intervals rather than as bare point
estimates, so that the strength of each claim can be judged against its
own uncertainty (Section~\ref{sec:limitations}).

Distractor noise creates a harder but still gradual source of degradation.
When the core chain is fixed and irrelevant rules are added, most models
remain reliable through moderate distractor counts, with failures appearing
mainly at higher values such as $D{=}32$, $D{=}64$, or $D{=}128$, depending
on the model. Larger models improve some cases, but the effect is not
uniform: reasoning-oriented models do not consistently outperform
instruction-tuned baselines on distractor resilience.

The clearest failure mode is contradiction detection. Instruction-tuned
models, including larger 24B--27B baselines, almost completely fail to
recognise inconsistent premise sets and usually continue to answer
according to the valid-looking chain. Reasoning-tuned models detect
contradictions more often, but they remain far below reliable accuracy and
show placement-sensitive behaviour. These results show an association between
inconsistent premises and low accuracy under the benchmark's abstention rule;
they do not by themselves isolate the cause of the observed errors.

These findings suggest that aggregate reasoning benchmarks may under-weight
one of the most diagnostic failure modes of current open-weight LLMs. A
model can appear reliable on long deductive chains while still confidently
deriving an answer from contradictory premises. This has direct
implications for retrieval-augmented generation, tool-using systems, legal
and scientific question answering, and other settings where conflicting
evidence is common.

\subsection{Scope and Limitations}
\label{sec:limitations}
We scope this study deliberately to keep the variable-isolation design
tractable on local hardware, and we state the resulting boundaries
explicitly rather than leaving them implicit.

\emph{Trial count and statistical power.} Each (phase, $v$, placement)
cell uses $n=20$ trials (40 for Phase~3), chosen to keep the full
sixteen-model $\times$ three-phase $\times$ eight-value sweep tractable
on an 8~GB and a 24~GB consumer GPU. As Table~\ref{tab:breakpoints} and
Figs.~\ref{fig:p1_all_models}--\ref{fig:p3_all_models} now show via
Wilson 95\% confidence intervals, this trial count leaves considerable
width around several point estimates (e.g.\ $\pm 0.15$--$0.20$ near
$p=0.5$), and the fixed $0.95$ breaking-point threshold is a point-estimate
convention rather than a statistically certified cutoff. We treat
individual accuracy values reported here as estimates with the stated
uncertainty, not exact quantities.

\emph{No frontier or closed-weight baseline.} All sixteen models are
open-weight and run locally; we do not evaluate closed frontier models
(e.g.\ GPT-4-class or Claude-class systems) because the benchmark targets
the local, resource-constrained deployment setting described in
Section~\ref{sec:intro}, and running such models under an identical,
auditable, temperature-0 protocol is not possible via typical hosted
APIs. Whether contradiction detection improves with frontier scale is
consequently left as the first item for future work below rather than
answered here.

\emph{Single decoding sample per condition.} All trials use greedy
decoding at temperature 0, so the 20 (or 40) trials per cell vary the
\emph{prompt instance} (via distinct random seeds) but not repeated
sampling of the same prompt. This isolates prompt-content variance
cleanly but does not characterise a model's run-to-run stability on a
fixed prompt, which would require repeated sampling at $T>0$ and is
outside the scope of this benchmark.

\emph{Early-stopping asymmetry.} As detailed in
Section~\ref{sec:earlystop}, the large instruction-tuned panel was run to
completion across the full sweep while the remaining panels used
early stopping; Section~\ref{sec:earlystop} gives the compute-budget
reason. This means Phase~3 distractor-scaling behaviour beyond $D=1$ is
characterised for only two models in this study.

\emph{Domain scope.} The benchmark isolates chain depth, distractor
volume, and contradiction placement using synthetic propositional
modus-ponens chains with non-semantic entity labels. This is a deliberate
design choice that removes world-knowledge and linguistic confounds, but
it also means the reported breaking points are properties of this
controlled setting and should not be read as direct predictions for
natural-language deductive tasks without further validation.

\emph{Answer labels and prompt interpretation.} The correct label is
\texttt{True} in Phases~1 and~2 and \texttt{Indeterminate} in Phase~3.
Consequently, phase comparisons can reflect answer preferences, task
interpretation, and differences in chain length as well as inconsistency.
Balanced answer labels and prompt variants are needed to distinguish these
explanations. Without those controls, the results cannot establish whether a
model internally detected a contradiction or understood the intended
abstention convention.

\emph{Distractor symbol collisions.} Distractor symbols are sampled in pairs
without a global uniqueness check against the core chain. A reconstruction
of the logged conditions found two Phase~3 conditions in the large
instruction-model sweep where a distractor shares a core-chain symbol
($D=32$, recency, trial 19; and $D=128$, recency, trial 6). The same core
contradiction remains present, but the distractors are not strictly
disjoint in these conditions. Results have not been recalculated after
excluding them.

\subsection{Future Work}
Future work should extend the benchmark to larger frontier models to test
whether contradiction detection improves with scale. It should also vary
the semantic distance between the contradiction and the chain conclusion,
examine how prompting and self-consistency affect inconsistency
recognition, replicate the benchmark across languages, increase per-cell
trial counts to tighten the confidence intervals reported in
Table~\ref{tab:breakpoints}, incorporate repeated sampling at $T>0$ to
characterise run-to-run stability, and analyse reasoning traces to
determine whether models internally notice contradictions but fail to
reflect that detection in the final answer.

% ----------------------------------------------------------------------
\section*{Declaration of competing interest}
The authors declare that they have no known competing financial interests
or personal relationships that could have appeared to influence the work
reported in this paper.

\section*{Funding}
This research received no funding.

\section*{Ethics statement}
This computational study did not involve human participants or animals. The authors confirm that institutional ethics approval was not required.

\section*{Data availability}
The benchmark prompt-generation code, model outputs, and analysis scripts
required to reproduce all reported results will be made publicly available
upon publication.

% ----------------------------------------------------------------------
\bibliographystyle{elsarticle-num-names}
\bibliography{references}

\end{document}